\documentclass[runningheads]{llncs}
\usepackage{graphicx}
\usepackage{color}
\usepackage[caption=false]{subfig}
\usepackage{float}
\usepackage{url}
\usepackage[acronym]{glossaries}
\begin{document}
\title{Optimized Wireless Control and Telemetry Network for Mobile Soccer Robots\thanks{Supported by RobôCIn, FACEPE and Centro de Informática - UFPE}}

\titlerunning{Optimized Wireless Control and Telemetry Network for Mobile S. Robots}%
%
%
\author{Lucas Cavalcanti \and
Riei Joaquim \and
Edna Barros}
\authorrunning{L. Cavalcanti et al.}
%
\institute{Centro de Informática, Universidade Federal de Pernambuco. \\
Av. Prof. Moraes Rego, 1235 - Cidade Universitária, Recife - Pernambuco, Brazil.
\email{\{lhcs, rjmr, ensb\}@cin.ufpe.br}}
\maketitle              
\newcommand{\fref}[1]{Fig.~\ref{#1}}
\newcommand{\tref}[1]{Table~\ref{#1}}
\newcommand{\crefrangeconjunction}{--}
\newcommand{\sref}[1]{Section~\ref{#1}}
\newcommand{\eref}[1]{Eq.~\ref{#1}}
\newcommand{\aref}[1]{Algorithm~\ref{#1}}

\newacronym{spi}{SPI}{Serial Peripheral Interface}
\newacronym{ssl}{SSL}{Small Size Soccer}
\newacronym{uart}{UART}{Universal Asynchronous Receiver / Transmitter}
\newacronym{rs232}{RS232}{Recommended Standard 232}
\newacronym{ttl}{TTL}{Transistor-Transistor Logic}
\newacronym{mac}{MAC}{Medium Access Control}
\newacronym{udp}{UDP}{User Datagram Protocol}
\newacronym{tcp}{TCP}{Transmission Control Protocol}
\newacronym{ieee}{IEEE}{Institute of Electrical and Electronics Engineers}
\newacronym{iot}{IoT}{Internet of Things}
\newacronym{crc}{CRC}{Cyclic Redundancy Check}
\newacronym{mbed}{mbedOS}{mbed Operating System}
\newacronym{csv}{CSV}{Comma-Separated Values}
\newacronym{cpu}{CPU}{Control Processing Unit}
\newacronym{larc}{LARC}{Latin American Robotics Competition}

\newacronym{wncs}{WNCS}{Wireless Networked Control Systems}

\begin{abstract}
In a diverse set of robotics applications, including RoboCup categories, mobile robots require control commands to interact with surrounding environment correctly. These control commands should come wirelessly to not interfere in robots' movement; also, the communication has a set of requirements, including low latency and consistent delivery. This paper presents a complete communication architecture consisting of computer communication with a base station, which transmits the data to robots and returns robots telemetry to the computer. With the proposed communication, it is possible to send messages in less than 4.5ms for six robots with telemetry enables in all of them.

\end{abstract}
\section{Introduction}
\label{sec:intro}



Competitions like RoboCup, the most significant autonomous robotics competition, creates high dynamic environments where robots should play soccer autonomously against another team \cite{Kitano:1997:RRW:267658.267738}. The soccer competition is perfect for developing interdisciplinary technologies, as it has obstacles and target positions are moving, similar to industries and warehouses. That dynamic requires accurate sensing of the field situation, combined with fast decision algorithms and real-time control.

With the evolution of wireless technologies, it became present in diverse solutions, including Industry 4.0 and smart houses. Moreover, in soccer categories, like \acrfull{ssl} \cite{rules:2019}, where a team of multiples robots plays soccer autonomously against another team, each team needs to communicate wirelessly with its robots. This communication is responsible for sending movements that the robot should follow on the field. 

At the \gls{ssl}, motion control is considered a system output where a complex robotic system is required. Furthermore, for wireless control robots, high accuracy and speed are needed. Then, communication is a critical factor to move the robots precisely. In the 2019 RoboCup, some \gls{ssl} teams had communication failures, which delayed its game. Also, there were teams with special requirements to avoid failures that affects the control of robots.

Besides the soccer robots, other applications, like industrial robots, require reliable wireless communication. Then, the main goal of this work is to propose an efficient wireless network for mobile robots. This work analyzes the network qualities through the application in \gls{ssl} competition, using RobôCIn's robots. For bringing the needed efficiency, the proposed network aims to reduce latency when sending control packets. It is achieved with a communication technology analysis, together with the embedded system design. This work goes more in-depth, building and analyzing peripherals connections, configurations, and proposing a protocol that minimizes data usage.

Considering that robots should be monitored, this work also brings a telemetry network without affecting the control latency to a point where control is compromised. Telemetry consists of sending information from robots, to the computer, giving the team's software the robot status. The telemetry network leverages the challenge because multiple robots should receive and send messages from one base station.

This work is divided as follows: Section 2 is a short introduction to wireless technologies and other team networks. In Section 3, the proposed system is described, together with the \gls{ssl} application description. In Section 4, the test method alongside the results is shown and discussed. Finally, in Section 5, the conclusion of this work is presented.


   


\section{Background and Related Work}
\label{sec:background}
This section goes through the technologies and characteristics necessary to build a radio frequency control network capable of controlling a mobile robot in real-time.

\subsection{Requirements of Wireless Communication} 
Wireless networks increase their usage, getting cheaper and better year over year. It is possible to see these technologies applied in \gls{iot} and Industry 4.0. Another application field affected by wireless technology is the \gls{wncs}\cite{WNDCS_ref}. It requires flexibility in wireless communication; however, the main focus is on controlling a remote system.

The system model with a \gls{wncs} has a higher risk because wireless may have bits failures, data loss, delay, and other problems due to environmental uncertainty and technologies. These problems increase the misbehaviors rate.

Besides the communication reliability, \gls{wncs} also requires low energy consumption to increase the system autonomy and high transference bandwidth to exchange the information needed \cite{WNDCS_ref}.

\subsection{Transmission Technology}
The technology for a wireless communication system needs to match the application requirements. The requirements matching reduces the risk of failure or malfunctioning. Several wireless technologies and protocols have been developed in the last years, and concepts like ubiquitous computing appeared.

There are several technologies and transmission protocols, and for building an optimized wireless control network, it is vital to analyze them and define which technology best matches the requirements.

In robot soccer, like \gls{ssl}, one concern is related to energy consumption since other components such as motors already generate great demand. Another factor is the time to deliver messages, which is essential to reduce errors in the control process. In addition, lower exposure to interference in the competition environment and a high data transmission rate are desired.

The nRF24l01 is a low energy consumption transmitter and developed, focusing on low latency communication, being advantageous concerning Wi-Fi \cite{WiFiAnalysis_ref}. In addition, Wi-Fi and Bluetooth networks are standard in every environment due to computers and smartphones. So, if one of these is adopted, we would have high interference. With the nRF24l01, the communication may operate at a frequency above Wi-Fi, Bluetooth, and ZigBee, reducing your exposure to interference \cite{nrf24l01}.

As for ZigBee \cite{NrfXBeeComp_ref}, its main disadvantage is its transmission rate, which is lower than the nRF24l01. Also, it has high exposure to interference because it uses the same frequency as Wi-Fi and Bluetooth. However, ZigBee has a lower transmission power than Wi-Fi, and Bluetooth \cite{ZigBeeAnalysis_ref}.

Modules like ZigBee and nRF24l01 were designed for embedded system. Then it is necessary to interface the cognition software that remains in a computer with an embedded system that transmits the data wirelessly. This embedded system is known as a radio base station, and the communication technology between computer and base station is also important to avoid bottlenecks.

A comparison of the main parameters of each technology is shown \tref{tab:SpecComp}, and, after analyzing the advantages and disadvantages of each technology, we chose the nRF24l01 developed by Nordic in 2008 \cite{nrf24l01}.

\begin{table}
\centering
\renewcommand{\arraystretch}{1}
\caption{The specifications of Bluetooth, ZigBee, WiFi and nRF24l01 modules\cite{OverviewOfWirelessTecnologies_ref}\cite{nrf24l01} \cite{NrfAutonomousRobots_ref}}
\label{tab:SpecComp}
\centering
\begin{tabular}{|l|c|c|c|c|}
\hline
Characteristics & Bluetooth & ZigBee &  WiFi  &  nRF24l01\\
\hline
{\bfseries Frequency} &  2.4 GHz & 0.86/0.91/2.4 GHz  & 2.4, 5 – 6 GHz & 2.4 - 2.48GHz. \\
{\bfseries Chanels} &  79 & 1/10/16 & 14  & 126\\ 
{\bfseries Speed Rate} &  3 Mbit/s & 20 - 250 Kb/s & 11 Mb/s - 10 Gb/s & 2 Mbps\\ 
{\bfseries Devices } &  8 & 65000 & 2007 & N/A\\ 
{\bfseries Radius} & 10 m & 10 - 100 m  & 100 m & 350 m \\
{\bfseries Power} &  0 to 30 dBm & 0 to 10 dBm & 15 to 20 dBm & -18 to 0 dBm\\ 
{\bfseries Current (TX, Stb)} &  40mA, 0.2mA & 30mA, 1$\mu$A & 400mA, 20mA & 13.5mA, 26$\mu$A\\
\hline
\end{tabular}
\end{table}

With that in mind, we chose to use an Ethernet connection, a standard available in some embedded systems and most computers. It is a standard that is easy to integrate and has data transmission that can reach up to 10Gbps. Different from Ethernet, the USB serial connection is common to debug and transfer the program between computers and boards, but is a slow protocol. And, due to the serial nature, it does not have support to continuous send and receive data at the same bus.

\subsection{Related Work}
Most wireless networks are modeled to \gls{iot} applications, where the main goal is to monitor. However, this work focuses on controlling mobile robots by using wireless networks. Then, the communication solution should minimize latency and provide reliability, which is fundamental for real-time applications, like mobile robots control.


There are studies developing robots that work with wireless networks \cite{NrfAutonomousRobots_ref}. The system's communication was built using nRF24l01 modules and analyzed the power consumption of each component. Although it succeeds in controlling an autonomous robot, it does not report any data from the communication efficiency and is tested in a single robot application.

Additional theoretical studies also gain relevance in network optimization, being essential to support the design decisions. In \cite{RobotSoccerComm_ref}, the author does not consider the nRF24l01 technology, but it concludes that radio configurations have a significant impact on the control network of soccer robots.

For narrowing the scope, some works report the impacts of nRF24l01 adoption in a wireless control network \cite{WirelessControlMobileDevices_ref}. The author discussed parameters, like data size and communication speed, to increase efficiency. Besides the parameters, the work reported tests with broadcast messages and multi-channel communication, but, in the end, did not report the latency of the whole wireless control network, crucial to the design of a robotic system.

Tigers Mannheim robotics team proposed a communication architecture for controlling its robots using the Ethernet protocol between the central unit and the base-station \cite{tiger2016}. The same work suggests using two nRF24l01 modules, one for sending messages and another for receiving messages. The Tigers team developed communication with real robots, but the team uses semi-autonomous navigation in their robots, which reduces the dependence on real-time communication to control the robots. Also, the Tigers' works did not perform communication or reliability tests.



\section{System Architecture}
\label{sec:architecture}
In this section, the wireless network is presented. It consists of a network that controls and monitors multiples mobile robots. It starts with commands from a computer connected to a base station, an embedded system that transmits and receives data through a wireless module. Each robot also has wireless modules responsible for receiving packets sent from the base station. For monitoring the robots, additional transceivers create a second channel of data exchange without interfering in the control messages transmissions.

There are two data exchange points to build the communication system: the first between the computer and the base station and the latter between the base station and robots. Both connections need to be optimized to achieve the lowest delay.

\subsection{Network Architecture}
The network implementation started by choosing the best modules that fit the system requirements. With that in mind, the technology that best fits is nRF24l01 \cite{nrf24l01}. In the proposed system, two modules in each embedded system are used to create the control and the telemetry communication. The nRF24l01 configuration and control interface is the Serial Peripheral Interface (SPI)\cite{SPIProtocol_ref}, which provides synchronized transmission and operates at a 10 Mbit/s transfer rate.

For the robot's microcontroller, the RobôCIn uses the NUCLEO-F767ZI \cite{f767zi} development board with an ARM Cortex M7 operating at 216 MHz, in the base-station, we chose the NUCLEO-H743ZI2\cite{h743zi}, an ARM Cortex M7 that operates at 400 MHz. This processor speed is essential to the base station since one of them communicates with multiple robots on the network.

For delivering a reliable communication system, it is necessary to develop robust software. For this, we use the environment of the \gls{mbed}\cite{mbed}, which is a real-time operating system for ARM micro-controllers that supports ARM peripherals and emulates virtual thread.

\begin{figure}
	\centering
	\includegraphics[width=0.88\linewidth]{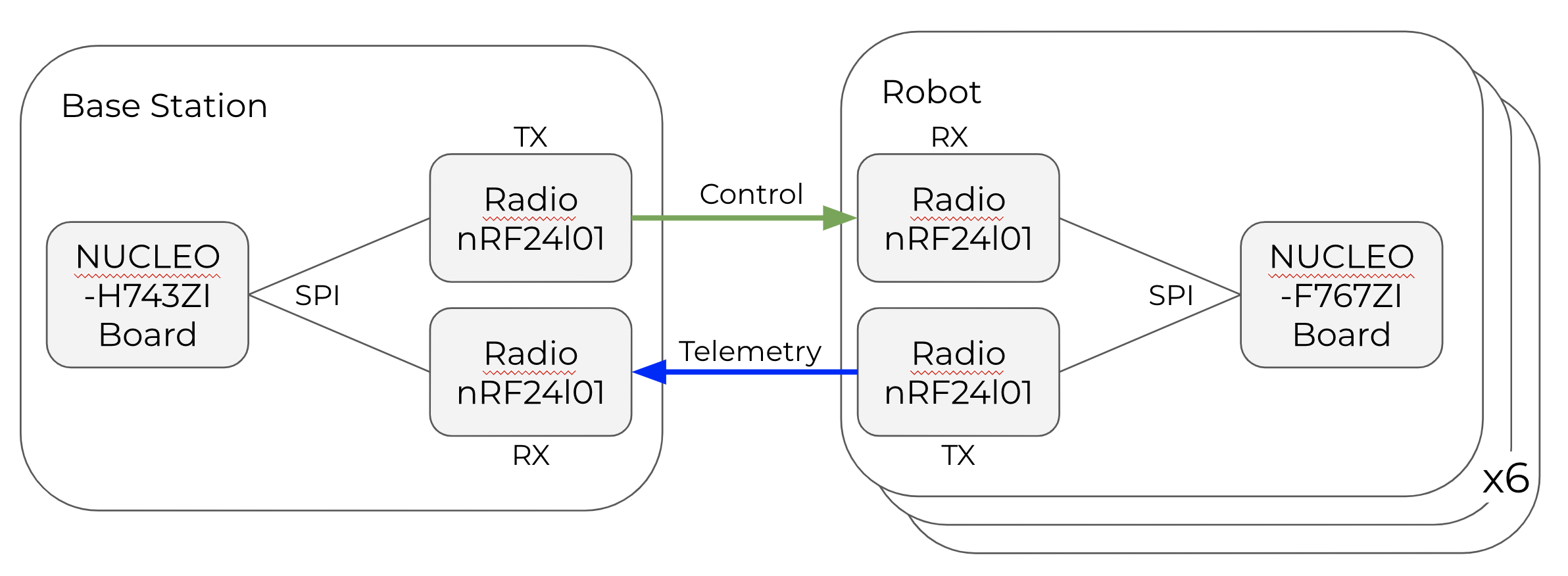}
	\caption{\textmd{Diagram of radios used for control and telemetry network on the Base Station and Robot, together with its way of communication.}}
	\label{fig:msg-exchange}
\end{figure}

The base-station, together with the proposed architecture, is shown in \fref{fig:msg-exchange}. The flow of control packets is made with one transceiver at the base station and another at the robot. This transceiver is configured as a sender radio to the robot's address in the base station, as shown in \tref{tab:projspec}. In the robots, it is in receiver mode with the same address.

\begin{table}[H]
\centering
\renewcommand{\arraystretch}{1}
\caption{Project Modules and Configuration in Base Station and Robots}
\label{tab:projspec}
\centering
\begin{tabular}{|c|c|c|}
\hline
 & {\bfseries Base Station} & {\bfseries Robots}\\
\hline
Transceiver &  nRF24l01 &  nRF24l01\\
Embedded Board &  Nucleo F767ZI & Nucleo H743ZI\\
CPU Frequency & 400 MHz & 216 MHz \\
Operating System & mbedOS & mbedOS \\
Radio Frequency (Control) & 2504 MHz & 2504 MHz \\
Radio Frequency (Telemetry) & 2529 MHz & 2529 MHz \\
Radio Address & 0x753FAD299ALL & 0x753FBD299ALL \\
\hline
\end{tabular}
\end{table}

The base station allows communication between the computer and the robots through Gigabit Ethernet. In the data connection, we use the \gls{udp} protocol to minimize the protocol overhead. As discussed previously, the Ethernet transfer delay is lower than Serial and the nRF24l01 transmissions, so the bottleneck stands in the wireless connection.

Finally, to build the telemetry in the network, an additional transceiver is used in the robot and the base station, as illustrated in \fref{fig:msg-exchange}. To report the robot's status to the computer, it checks how long it has sent telemetry; when it is longer than the configured value, it measures the robot's status and sends the telemetry encoded back to the base station. The transceivers in the robot are configured in sender mode and address the base station (\tref{tab:projspec}).

For receiving a telemetry packet, the base station has a virtual thread, parallel to the control code, to deal with the incoming messages. The second transceiver in the base station is configured as a receiver, expecting messages at the telemetry frequency. So, whenever a message reaches the telemetry transceiver, the base station forwards to the computer, where the message is appropriately decoded and identified.

\subsection{Communication Protocol}
After defining the system's architecture and its technology, it was essential to determine the communication topology to minimize the configuration overhead, the dependency of message order, and the packets' loss. When considering the control and telemetry messages, there is no need to recover old packets. So, the topology chosen was the star network, in which the base station sends and receives messages of all robots. There is no overhead of dynamic configuring the parameters of the transceiver with this topology, and no acknowledge packet is necessary. On the other hand, the message should include a robot identification so that each robot can identify its message.

Finally, with the chosen communication architecture and topology,  a message protocol has been developed for control messages. It aims to minimize the payload size to optimize the delivery time. So, in the first half byte, there is the message type identification, at the second half, there is the robot identifier. These initial bytes enable the robot to recognize different messages and filters the ones addressed to them. Additional bytes define the robot's movements, like linear and angular speed, and its peripheral's actions.

For telemetry, another messaging protocol has been created, but equally to the control protocol, the telemetry message has the message and robot identification in the first byte. But, differently from the control protocol, the telemetry protocol has each motor speed, measured by the robot's sensors, battery level, and peripherals state.

No matter the message, the base station only needs to re-transmit messages. Then, the protocol encodes and decodes messages for computers and robots only. The data exchange goes by Ethernet, between computer and base-station, and nRF24l01, base-station, and robot.

\subsection{Applying to Small Size League (SSL)}
This section presents how the proposed communication has been used in the \gls{ssl} competition. So, it was necessary to build the control packet protocol, including the information required to control the robot, and monitor it, create the telemetry packet. In the following, the control and telemetry packets defined are presented.

\textbf{SSL Control Packet}: Message Type(4 bits), Robot ID(4 bits), Vx - Linear Speed(20 bits), Vy - Linear Speed(20 bits), $\omega$ - Angular Speed(20 bits), $\theta$ - Robot Angle(20 bits), Kick Front(1 bits), Kick Chip(1 bits), Charge the Kick(1 bits), Strength of the Kick(8 bits), Turn on the Dribbler(1 bits), Speed of the Dribbler(8 bits), Additional Command(4 bits). With a total of 14 bytes of payload.

\textbf{SSL Telemetry Packet}: Message Type(4 bits), Robot ID(4 bits), m1 - Motor 1 Speed(16 bits), m2 - Motor 2 Speed(16 bits), m3 - Motor 3 Speed(16 bits), m4 - Motor 4 Speed(16 bits), Dribbler’s Motor Speed(15 bits), Kick’s Capacitor Load(8 bits), Ball on the Robot(1 bits), Robot’s Battery(8 bits). With a total of 13 bytes of payload.

Once both communication packets were defined, the robot and computer can understand each other the messages. The next step is configuring the base station and robot transceivers. One address and frequency channel for a pair of transceivers are necessary, one in the base station and another in the robot.

\section{Validation and Results}
In this section, the proposed base-station is analyzed in the \gls{ssl} environment. Based on these results, the parameters that optimize the communication were found.

\subsection{Time Analyses}
The interval between each message transmission is essential and different for each communication system. It is necessary because computers are normally faster than embedded systems. Then, an uninterrupted flow of messages may increase the overhead of the base station. Moreover, this work searches the ideal interval period based on the delivery delay at the robots.

The delivery time test measures the interval between messages that should arrive at a robot. The interval between messages varies because the network works asynchronously and may lose packages. So, here, the delivery time between messages means the average interval between 500 messages received. Due to the reception variation between messages, the standard deviation is also calculated.

The test flow begins with the computer sending messages to robots via the base station, with some send interval. The robot measures the interval between 500 messages separately. After the measurements, the robot reports each interval to a computer, where the data is analyzed. With the results analyzed, another interval is configured at the computer to search for the optimal one.

Although the telemetry impacts the communication delivery performance, the test uses the same flow to test the telemetry. So, the optimal interval for control packets is configured in the computer.  Another interval is configured at the robots for the telemetry. The results reveal the impact of telemetry in the control network.

\subsubsection{Control Message Interval Time}
\begin{figure}
    \centering
    \captionsetup[subfloat]{justification=centering}
    \subfloat[Using Serial station]{\includegraphics[width=.45\linewidth, scale=1]{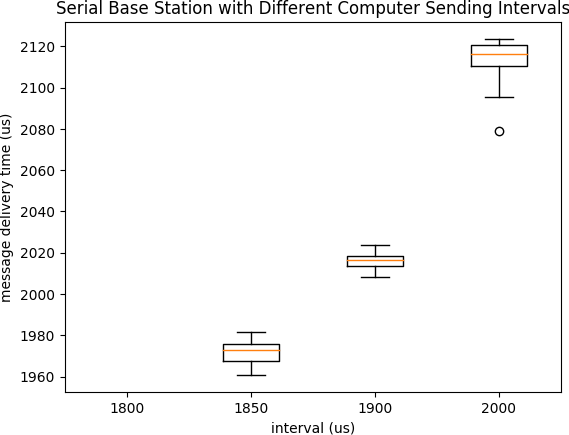}}
    \label{fig:serial-interval}
    \subfloat[Using Ethernet station]{\includegraphics[width=.45\linewidth, scale=1]{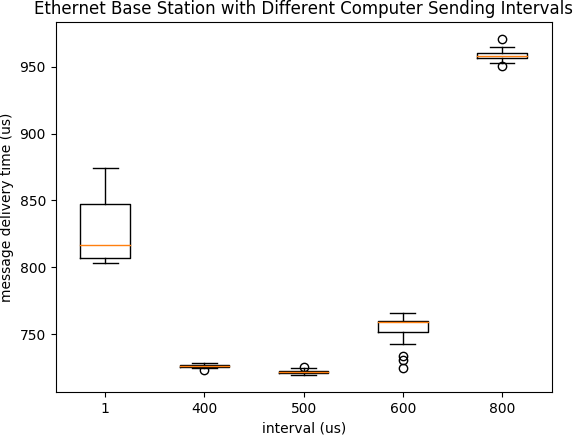}}
    \label{fig:eth-interval}
    
	\caption{\textmd{Reception delay for 25 tests at each different sending interval, using Serial(a) or Ethernet(b) base station transmitting computer messages at the configured sending intervals}.}
	\label{fig:interval}
\end{figure}

Analyzing the results in \fref{fig:interval}(a), the base-station using serial interface has a fast and reliable throughput at 1900$\mu$s of an interval between each message sent. Smaller intervals caused a bit flip in communication, which causes undesired robot behavior.

The \fref{fig:interval}(b) shows the test result with an optimum interval time for Ethernet interface is 500us. Almost four times smaller than the Serial, the Ethernet approach does not corrupt the bits with a shorter interval time than 500us but increases the delivery time.

\subsubsection{Telemetry Message Interval Time}
Finally, the telemetry impact in the control network is analyzed. The control messages interval was measured in a scenario with no telemetry and an unknown telemetry interval. Again, the tests were applied to find an optimal interval for sending telemetry packets. The test analyzes the delivery time of control packets, as it is essential to robots even with telemetry.

The disadvantages of the serial interface and the results above lead to discarding the serial interface to control communication with telemetry.  After all, the tests initially used the interval of 200ms between telemetry messages because of RobôCIn's requirement. The tests decreased the telemetry interval to find a balance that guarantees quickly sampling without damaging the control network.

\tref{tab:telemetry} shows the network time performance of an Ethernet base station, sending control packets with 500 microseconds of interval. The test applied different telemetry intervals, so, \tref{tab:telemetry} presents an impact of 0.39\% with a 200ms telemetry interval; at 50ms of sampling, the reception interval increases 1.23\%. Furthermore, the 10ms sampling increases 6.48\% of average delivery time. Then, the Ethernet base station receiving telemetry packets at every 50 milliseconds guarantees a control update every 730.89 microseconds.

\subsection{System Validation}
The network efficiency was tested and validated by changing the number of robots, its distances to the base station, and enabling the telemetry. These tests were performed in the RobôCIn field, focusing on the Ethernet configuration and serial interface.

\subsubsection{Different Distances}
An important characteristic is a robustness in different distances. For simulating that environment, the robot was, first, positioned 0.4m from the transceiver, after it was 2.5m of distance, and finally at the opposite side of the field, 5m of distance.

\begin{figure}
    \centering
    \captionsetup[subfloat]{justification=centering}
    \subfloat[Delivery time of Serial base station different distances]{\includegraphics[width=.44\linewidth, scale=1]{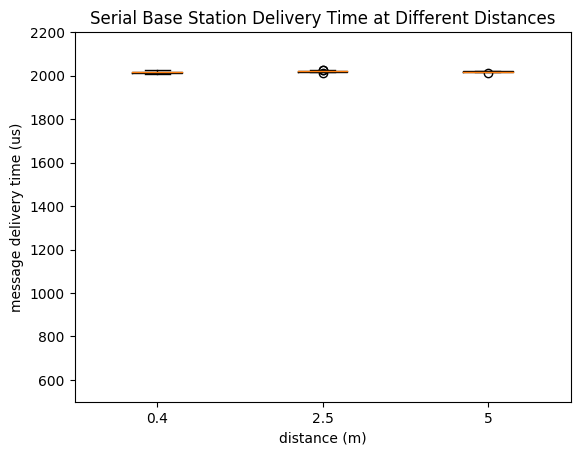}}
    \label{fig:distance-ser}
    \subfloat[Delivery time of Ethernet base station different distances]{\includegraphics[width=.44\linewidth, scale=1]{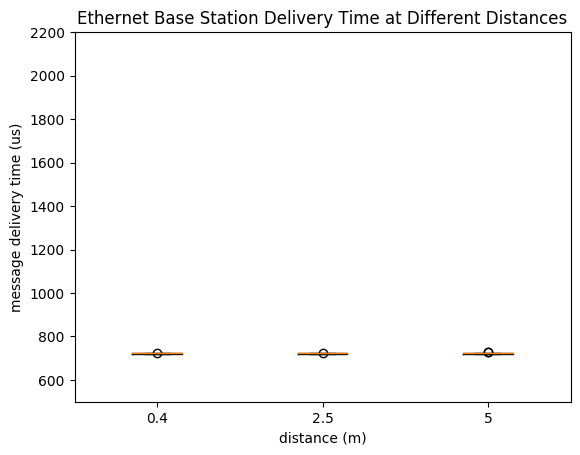}}
    \label{fig:distance-eth}
	\caption{\textmd{Delivery time of base station with different distance between transceivers}.}
	\label{fig:distance}
\end{figure}

In \fref{fig:distance} shows a consistent delivery time in robots with previous results; the differences appear only in the communication interface.

\subsubsection{Multiple Robots}
Today, the \gls{ssl} there are six robots from each team in Division B and 11 in Division A. So, the network needs to communicate with multiple robots.

\begin{figure}[H]
	\centering
	\captionsetup[subfloat]{justification=centering}
    \subfloat[Delivery time with different number of nodes at Ethernet base station]{\includegraphics[width=.44\linewidth, scale=1]{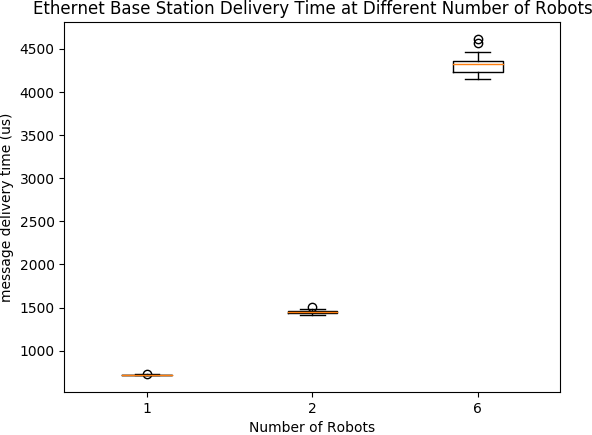}}
    \label{fig:mult-r-eth-1-6}
    \subfloat[Delivery time at 6 robots network with Ethernet base-station from two different robots \label{fig:mult-comp-6r}]{\includegraphics[width=.44\linewidth, scale=1]{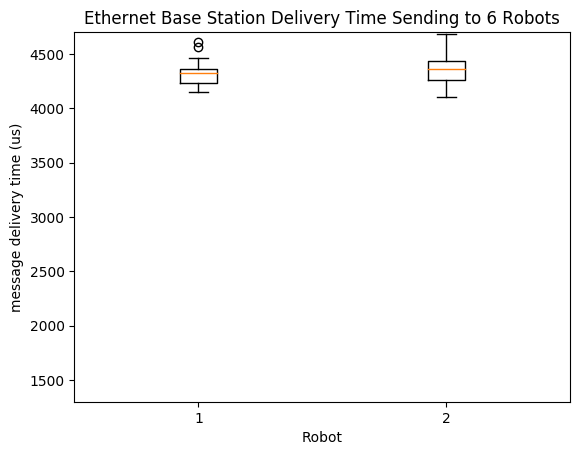}}
	\label{fig:multp-compare}
	\caption{\textmd{Delivery time analysis between network size (1, 2 and 6 robots)}.}
	\label{fig:mult-robot}
\end{figure}

After testing with 1, 2, and 6 robots using the Ethernet interface, the results, available in \fref{fig:mult-robot}(a), revealed that the delivery time is proportional with the number of robots. Outcome expected because the base station sends six messages to control six robots, one for each robot.

Therefore, to test consistency, tests were realized at two different robots separately. The test result, presented in \fref{fig:mult-robot}(b), confirms the expectation that a network with six robots has a delivery time similar for two different robots. Even though each robot receives all messages, the test considers only the packets addressed to the given robot.

\subsubsection{Telemetry}
To safely use telemetry in games, the control network was tested with telemetry enabled for six robots. The telemetry interferes in base-station flow; then, it may interfere with the control delay. After testing, the results, presented in \tref{tab:telemetry}, show an increase of tens of microseconds. And, using a telemetry sampling time of 50ms, for each robot, the delay increase is less than 2\%. This result does not compromise the control efficiency.

\begin{table}
\centering
\renewcommand{\arraystretch}{1.2}
\caption{Ethernet Base Station with Telemetry Different Sampling Interval}
\label{tab:telemetry}
\centering
\begin{tabular}{|c|cc|cc|}
\hline
\bfseries Test Condition &\multicolumn{2}{|c|}{\bfseries Delivery Time} & \multicolumn{2}{|c|}{\bfseries Increase}\\
& Average & Standard Deviation & Average & Standard Deviation\\ 
\hline
Without Telemetry & 721.98$\mu$s & 140.12 & &	\\
200ms Sampling & 724.78$\mu$s & 159.44	& 0.39\% & 13.79\% \\
50ms Sampling & 730.89$\mu$s & 196.07	& 1.23\% & 39.93\% \\
10ms Sampling & 768.80$\mu$s & 217.69	& 6.48\% & 55.36\% \\
\hline
\end{tabular}
\end{table}
        

\section{Conclusion}
This paper presented a wireless network architecture to control mobile robots. The communication built was applied and validated in RobôCIn \gls{ssl} robots, using the competition environment as the system requirement. The soccer competition simulates a dynamic environment, where all the obstacles and goals are moving. Then, the control system needs a quick reaction and good reliability to change robots' movements.

Usually, communication is not the focus of the control system or soccer teams in RoboCup; however, it is essential to accomplish a high rate of delivered messages. Then, this work brings a simple communication system that leverages the control and monitor capabilities of teams and wireless systems. The work presents the modules, their connections, and required configuration to work inside the proposed architecture.

In the paper, the communication validation uses the delivery time of messages in the robots, as it is the main objective of a wireless control system. The results show an average delivery time smaller than one millisecond, for one robot, even with robots telemetry. Moreover, although the delivery time increases with the number of robots, the proposed communication achieves a \gls{ssl} team, a delivery time smaller than 16 milliseconds, a typical specification of competition cameras. Future works will analyze energy consumption and bandwidth in each component to evaluate possible optimization opportunities. Moreover, future work will introduce inter robots communication and analyze new communication modules with different technologies.
\subsubsection*{Acknowledgements}
The  authors  would  like  to  acknowledge  the  RoboCIn's team  and  Centro  de  Informática  -  UFPE  for  all  the  support  in  this research. The authors would like to thanks the founding provided by FACEPE (Fundação de Amparo a Ciência e Tecnologia de Pernambuco).

\bibliographystyle{splncs04}
\bibliography{default_content/references}

\end{document}